\documentclass[letterpaper, 10 pt, conference]{ieeeconf}  
\usepackage{graphicx}
\usepackage{color}
\usepackage{booktabs}
\usepackage{stfloats}
\usepackage{multirow}
\usepackage{cite}

\IEEEoverridecommandlockouts                              

\title{\LARGE \bf
FEANet: Feature-Enhanced Attention Network for RGB-Thermal Real-time Semantic Segmentation
}

\author{Fuqin Deng$^{1,3,4\ast}$, Hua Feng$^{1\ast}$, Mingjian Liang$^{1}$, Hongmin Wang$^{1}$, Yong Yang$^{4}$, Yuan Gao$^{3}$, \\Junfeng Chen$^{3}$, Junjie Hu$^{3}$, Xiyue Guo$^{3}$, and Tin Lun Lam$^{2,3,\dagger}$
\thanks{This work was supported in part by the National Key R$\&$D Program of China (2020YFB1313300), the funding (AC01202101025, AC01202101026) from the Shenzhen Institute of Artificial Intelligence and Robotics for Society, the special projects in key fields of Guangdong Provincial Department of Education (2019KZDZX1025), Innovative Program for Graduate Education (503170060259) from the Wuyi University, and Shenzhen Peacock Plan of Shenzhen Science and Technology Program (KQTD2016113010470345).
We would also like to thank Dr. Li Nan Nan from Macau University of Science and Technology for his valuable discussion on this paper.}
\thanks{$^{\ast}$Authors contributed equally}
\thanks{$^{1}$School of Intelligent Manufacturing, the Wuyi University, Jiangmen, China.
        }%
\thanks{$^{2}$School of Science and Engineering, the Chinese University of Hong Kong, Shenzhen, China.
        }%
\thanks{$^{3}$Shenzhen Institute of Artificial Intelligence and Robotics for Society, the Chinese University of Hong Kong, Shenzhen, China.
        }%
\thanks{$^{4}$3irobotix Co.,Ltd, Shenzhen, China.
        }%
\thanks{$^{\dagger}$Corresponding author is Tin Lun Lam
        {\tt\small tllam@cuhk.edu.cn}
        }%
}

\begin{document}

\maketitle
\thispagestyle{empty}
\pagestyle{empty}

\begin{abstract}

The RGB-Thermal (RGB-T) information for semantic segmentation has been extensively explored in recent years. However, most existing RGB-T semantic segmentation usually compromises spatial resolution to achieve real-time inference speed, which leads to poor performance. To better extract detail spatial information, we propose a two-stage Feature-Enhanced Attention Network (FEANet) for the RGB-T semantic segmentation task. Specifically, we introduce a Feature-Enhanced Attention Module (FEAM) to excavate and enhance multi-level features from both the channel and spatial views. Benefited from the proposed FEAM module, our FEANet can preserve the spatial information and shift more attention to high-resolution features from the fused RGB-T images. Extensive experiments on the urban scene dataset demonstrate that our FEANet outperforms other state-of-the-art (SOTA) RGB-T methods in terms of objective metrics and subjective visual comparison (+2.6\% in global mAcc and +0.8\% in global mIoU). For the 480 × 640 RGB-T test images, our FEANet can run with a real-time speed on an NVIDIA GeForce RTX 2080 Ti card.

\end{abstract}

\section{INTRODUCTION}

As a fundamental but challenging task in computer vision, semantic segmentation can be broadly applied to the fields of path planning, autonomous driving, and video surveillance \cite{kang2011multiband} \cite{li2017traffic} \cite{chen2018importance}. Most of the existing deep learning-based semantic segmentation networks \cite{garcia2017review} \cite{romera2017erfnet} mainly deal with RGB images. However, RGB images could provide less information for the model training and produce inaccurate prediction results on the scenarios of similar texture, complex background with dim light, or total darkness. With the popularity of thermal imaging cameras, some
\begin{figure}[h]
    \centering
    \includegraphics[width=3.4in]{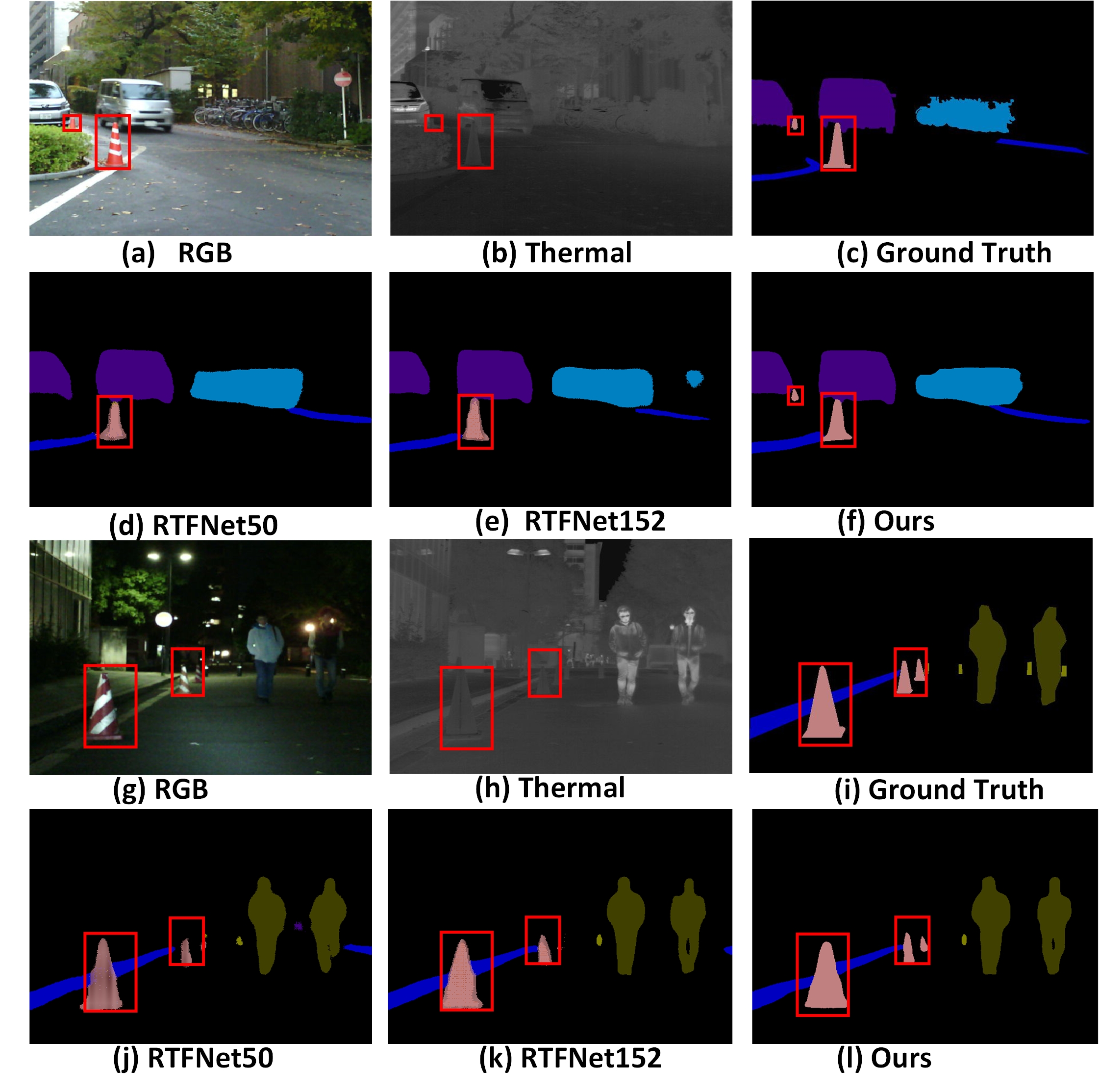}
    \vspace{-5mm}
    \caption{Qualitative comparison with two latest networks in daytime and nighttime urban street scenes. The color cones (the objects marked in the red frame) are too small to detect and segment. We can see that the color cone boundaries segmented by the RTFNet are not so sharp or fail to segment them correctly (e.g., (d), (e), (j), (k)), whereas our FEANet gives a more ideal segmentation result (e.g., (f), (l)).}
    \label{fig:fig1}
    \vspace{-5mm}
\end{figure}
researchers found that thermal information is robust and effective for reducing the ambiguity in challenging lighting conditions \cite{ha2017mfnet}, such as in urban street scenes. Therefore, the thermal images created by the thermal imaging cameras can be added as important supplements to improve the performance of RGB-T semantic segmentation.

Recently, RGB-T semantic segmentation has received increasing research attention. Various RGB-T models \cite{sun2019rtfnet} \cite{shivakumar2020pst900} \cite{sun2020fuseseg} have been proposed to improve the segmentation performance by combining RGB and thermal information. However, the performance of existing RGB-T models tends to drastically decrease when faced with certain complex scenarios (e.g., cluttered backgrounds, varying illuminations). Therefore, the existing RGB-T methods still need to solve the following challenges in order to make further progress.

The first challenge is to effectively extract multi-level features from RGB-T fused data. Generally, high-level features contain rich semantic information which can be used for object location, while low-level features provide plentiful micro details that are useful for reducing glitch noise and refining segmentation boundaries. Therefore, current RGB-T semantic segmentation methods (e.g., MFNet \cite{ha2017mfnet}, RTFNet \cite{sun2019rtfnet}) use either a direct feature extracting strategy or a progressive multi-data fusing process to leverage multi-level features. However, due to the direct multi-level features extracting and merging strategy without considering differences between levels, these processes suffer from the incomplete extraction problem of noisy low-level features. As shown in Fig.~\ref{fig:fig1}, the segmented object boundaries in the RTFNet prediction maps are not sharp (e.g., (d), (e), (j), (k)).

The second challenge is to excavate informative features from the thermal modality. Thermal images are of low quality, which leads to unpredictable noise during the data fusion process. Previous RGB-T models usually treat the extra thermal images as a fourth-channel input without the modification of three-channel RGB encoder stream or fuse RGB and thermal features by simple summation and multiplication. These methods treat thermal and RGB information from the same perspective and ignore the fact that RGB images contain color and texture, whereas thermal maps contain the spatial relations among objects. Due to this modality difference, the above-mentioned simple combination methods \cite{sun2019rtfnet} \cite{sun2020fuseseg} are not effective. As shown in Fig.~\ref{fig:fig1} (d), (e), (j), (k), the RTFNet fails to detect and segment the small target objects (e.g., color cones).

To address the above issues, we propose a two-stage FEANet for better RGB-T semantic segmentation performance. As shown in Fig.~\ref{fig:fig2}, our FEANet contains a two-stages process for feature extraction and fusion. In stage 1, we introduce a FEAM module, which exploits the inter-channel and spatial relations for the detail information. The proposed FEAM exploits multi-level features in a progressive refinement way to suppress distractors in the encoder stream. This strategy is based on the observation that low-level features provide discriminative semantic information with micro details, which may contribute significantly to eliminating the background distractors. In stage 2, to improve the compatibility of RGB and thermal features, the corresponding RGB and thermal feature maps are aggregated through elementwise summation into the RGB encoder stream. Our main contributions are summarized as follows:

\begin{itemize}

\item We design a two-stage FEANet to deal with the object boundaries and the small target object for RGB-T semantic segmentation in urban scenes.
\item We introduce a FEAM module to enhance multi-level features and fuse RGB and thermal information in a complementary way.

\end{itemize}

The remainder of this letter is structured as follows. In section II, related works have been reviewed. In section III, we describe our network in detail. In section IV, experimental results and discussions are presented. Conclusions and future work are drawn in the last section.

\section{RELATED WORKS}

\subsection{Semantic Segmentation}

Over the past few years, semantic segmentation is a great challenge for detecting and locating target objects in computer vision. The Convolutional Neural Networks (CNNs) \cite{krizhevsky2012imagenet} had been applied to improve the accuracy in the image classification and semantic segmentation tasks since 2012. In 2015, Fully Convolutional Networks (FCN) \cite{long2015fully} had been proposed for the semantic segmentation, which had an end-to-end network architecture and outperformed the traditional methods that rely on the hand-crafted features extraction mechanism. 

Similar to the FCN, the SegNet \cite{badrinarayanan2017segnet} adopted the Encoder-Decoder network architecture as the backbone for semantic segmentation tasks. The SegNet used a pre-trained VGG16 architecture as its encoder, and then applied the output of the encoder as the input of the up-sampling decoder. SegNet achieved state-of-the-art accuracy while getting the low inference speed. In subsequent years, the Encoder-Decoder network structure is widely used in semantic segmentation methods. UNet \cite{ronneberger2015u} and DeepLabv3 \cite{chen2018encoder} have the large Encoder-Decoder architecture, specially, the decoder can restore the high-resolution feature maps from the low layers through the short-cut connections.

Real-time semantic segmentation methods aim to generate high-quality segmentation results in real-time. ENet \cite{paszke2016enet} also followed the Encoder-Decoder architecture to achieve real-time semantic segmentation, but it was optimized for fast inference and high accuracy. Due to the efficiency of ENet, it can effectively process images (480×640 RGB) in the requiring high-speed inference situations. However, ENet failed to perform as well as SegNet on spectral images datasets, such as SUN RGB-D \cite{song2015sun}.

To improve the accuracy and speed of semantic segmentation, the BiSeNet \cite{yu2018bisenet} had the spatial path that preserves the spatial information and the semantic path that obtains the sufficient receptive field. Based on these two paths, a new Feature Fusion Module was developed to combine the features efficiently. However, the BiSeNet just captures the information from the lower layer to sharpen the boundaries with a slow inference speed.

\subsection{RGB-T Semantic Segmentation}

Some methods adopting CNNs were designed for the RGB-D dataset, which contains images that were acquired by multispectral cameras. In these works, we found that some ideas were useful for designing our method. Hazirbas et al. \cite{hazirbas2016fusenet} proposed a new CNN network named FuseNet, which contained an Encoder-Decoder structure that simultaneously extracts features from RGB and depth images. In \cite{wang2018depth}, RGB and spectral images feature maps were not only processed separately in the encoder stream but also in the decoder stream.

The existing urban scenario image segmentation datasets are based on visible spectral images (RGB images), such as Cityscapes \cite{cordts2016cityscapes} and Daimler Urban dataset \cite{scharwachter2013efficient}. Naturally, semantic segmentation methods based on these datasets can only be used to process RGB images. Furthermore, most of these methods focused only on improving the segmentation accuracy while neglecting the inference speed. For RGB-T semantic segmentation of urban scenes, MFNet, RTFNet, and FuseSeg-161 were proposed to fuse RGB and thermal data in a novel Encoder–Decoder structure. In this structure, two identical encoders were employed to extract features from RGB and thermal data, respectively, and one decoder was designed to gradually restore the resolution. In addition to the above methods, recently, other RGB-T fusion methods \cite{stone2021deepfusenet} \cite{jayasuriya2020active} utilized the combination of omnidirectional (O-D) infrared sensors and O-D visual RGB sensor for semantic segmentation in autonomous robotic systems.

\begin{figure*}[ht]
    \centering
    \includegraphics [width=6.2in]{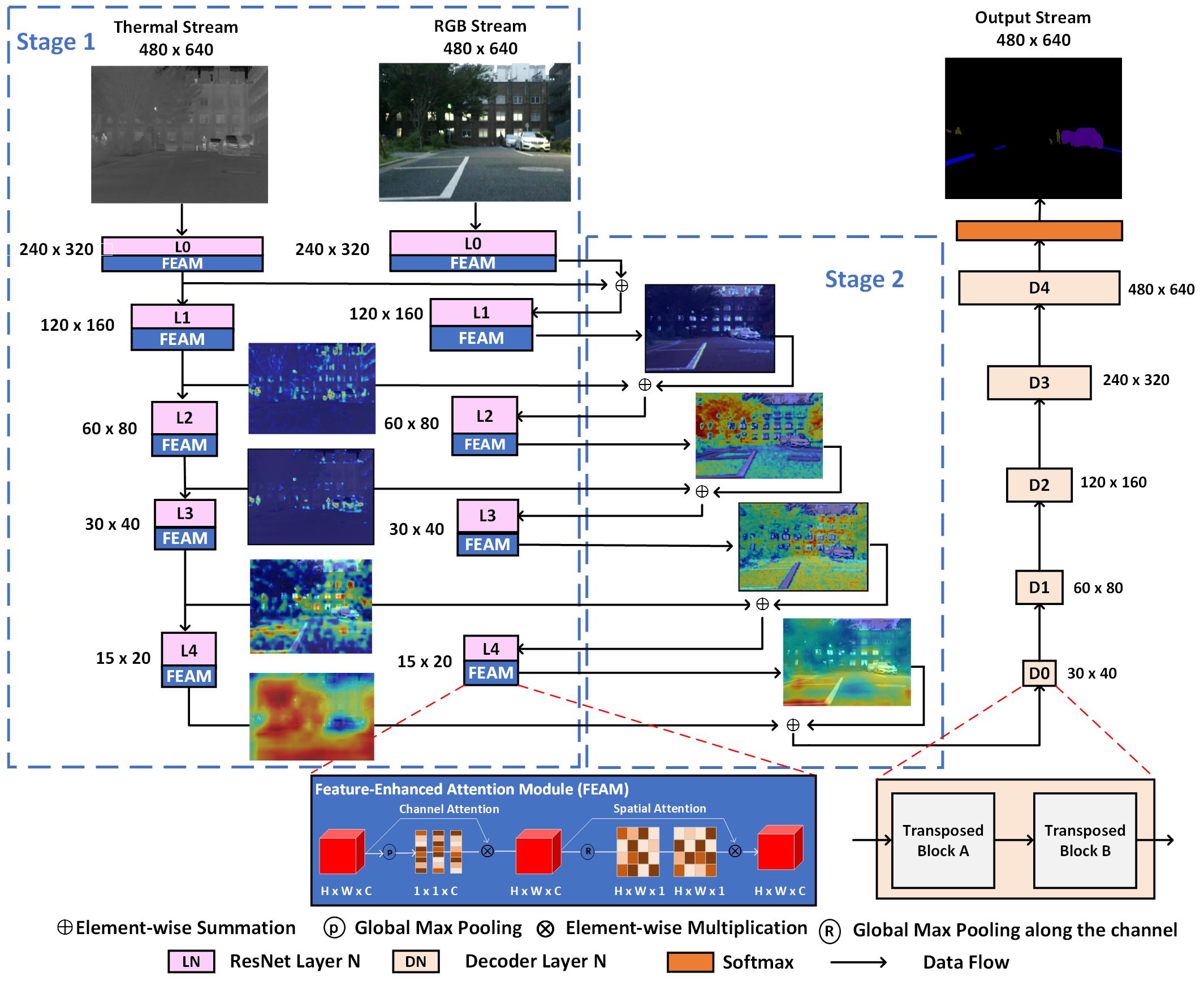}
    \caption{The overall architecture of the proposed FEANet. From left to right are Thermal Stream, RGB Stream, and Output Stream. The encoder in Thermal Stream and RGB Stream contains two extracting stages. In stage 1, Thermal Stream and RGB Stream use ResNet \cite{he2016deep} as the feature extractor layer. The output part of each layer is weighted through the FEAM. In stage 2, the output map of Thermal Stream is fused into the RGB Stream. The decoder in Output Stream is composed of Transposed blocks A and B.}
    \label{fig:fig2}
    \vspace{-3mm}
\end{figure*}


\section{PROPOSED METHOD}
In this section, we first introduce the overall architecture of our FEANet, which contains two extracting encoder streams and an output decoder stream. As the Encoder-Decoder structure has been confirmed as an effective architecture in many semantic segmentation networks, our FEANet also adopts this structure. Then, to fully excavate informative cues from both the RGB and thermal feature maps, we present the FEAM to enhance the multi-level features for superior segmentation performance.

\subsection{Overall Architecture} 

As shown in Fig.~\ref{fig:fig2}, our FEANet contains two main steps: feature extracting and resolution restoring. Two encoder streams and one decoder stream are designed for the feature extraction and recovery, respectively. In the feature extraction progress, two encoders extract the multi-level features from three-channel RGB and one-channel thermal images, respectively. With increasing encoder stream depth, the high-level features (e.g., L3, L4 in Fig.~\ref{fig:fig2}) will be more useful for capturing global context, while they lose the object details. When we up-sample the high-level feature maps, the output prediction will be blurred and object boundaries will become unclear. Instead, the proposed FEAM can distinguish object regions which are too small to detect. In the resolution restoration progress, the decoder gets dense output predictions. At the end of FEANet, the final softmax layer is adapted to get the prediction output map for the RGB-T semantic segmentation results.

\begin{figure}[t]
    \centering
    \includegraphics[width=3.4in]{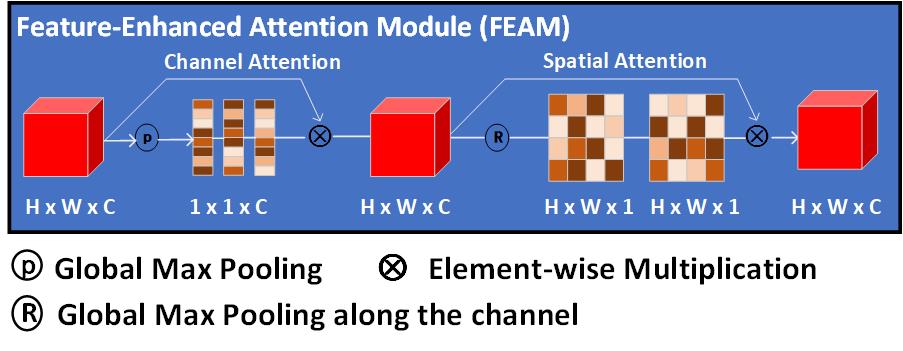}
    \vspace{-5mm}
    \caption{Architecture of the Feature-Enhanced Attention Module (FEAM)}
    \label{fig:fig3}
\end{figure}

\begin{figure}[t]
    \centering
    \includegraphics[width=3.2in]{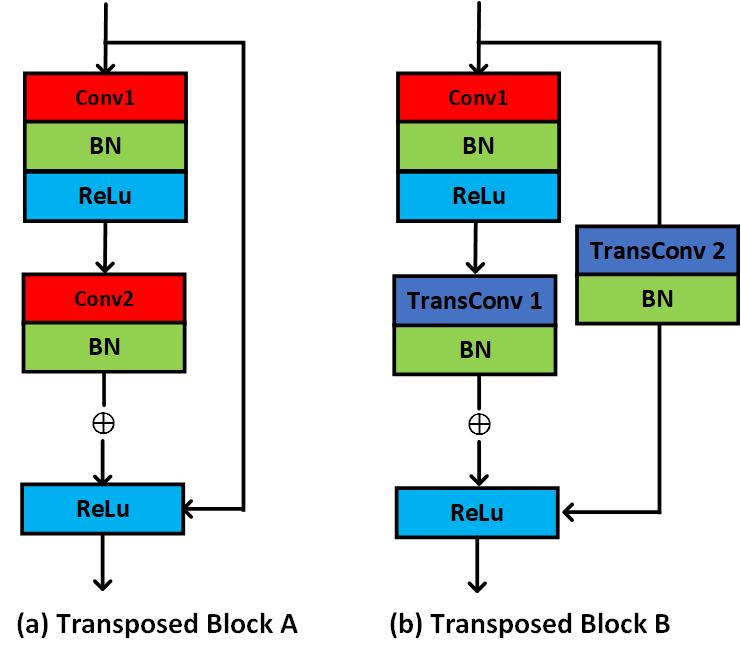}
    \caption{ Architecture of the Transposed block. Conv, TransConv and BN refer to the convolutional layers, transposed convolutional layers and the batch normalization layers, respectively. The detailed channel numbers for Conv and TransConv layers are listed in Tab.~\ref{tab:tab1}.}
    \label{fig:fig4}
    \vspace{-5mm}
\end{figure}

The proposed FEANet explores the two-stages cross-modal fusion methods. In the first stage, we first extract both the RGB and thermal feature maps by the ResNet block and then refine the detail features through the FEAM module. In the second stage, the corresponding RGB and thermal feature maps are aggregated through elementwise summation into the RGB encoder stream. At the end of two encoders, the final refined feature maps are transmitted to the decoder. With the two-stage feature extracting strategy, the loss of rich semantic information through the intensive feature extracting could be recovered.

\subsection{Encoder-Feature Extracting}

In the proposed FEANet, both the RGB and thermal features are extracted in two encoder streams. Specifically, both the RGB and thermal encoder streams employ five convolutional blocks from ResNet \cite{he2016deep} as the standard backbone and attach an additional FEAM after every single convolutional block, respectively. Usually, the existing ResNet is designed for the three-channel RGB images extracting, which is not suitable for the single-channel images, then we modify the number of the first convolutional layer to be one to extend it to the thermal image.

\begin{table}[htbp]
  \centering
  \caption{Configuration for the convolution (Conv) and transposed convolution (TransConv) layers in the individual module of the decoder.}
    \begin{tabular}{rrrcccccccc}
    \toprule
          &       & \multicolumn{2}{c}{Name} & \multicolumn{2}{c}{Kernel Size} & \multicolumn{2}{c}{Stride} & \multicolumn{2}{c}{Padding} \\
    \midrule
    \multicolumn{2}{c}{\multirow{2}*{Block A}} & \multicolumn{2}{c}{Conv 1} & \multicolumn{2}{c}{3x3} & \multicolumn{2}{c}{1} & \multicolumn{2}{c}{1} \\
    \cmidrule{3-10}
          &       & \multicolumn{2}{c}{Conv 2} & \multicolumn{2}{c}{3x3} & \multicolumn{2}{c}{1} & \multicolumn{2}{c}{1} \\
    \midrule
    \multicolumn{2}{c}{\multirow{4}*{Block B}} & \multicolumn{2}{c}{Conv 1} & \multicolumn{2}{c}{3x3} & \multicolumn{2}{c}{1} & \multicolumn{2}{c}{1} \\
    \cmidrule{3-10}
          &       & \multicolumn{2}{c}{TransConv 1} & \multicolumn{2}{c}{2x2} & \multicolumn{2}{c}{2} & \multicolumn{2}{c}{0} \\
    \cmidrule{3-10}
          &       & \multicolumn{2}{c}{TransConv 2} & \multicolumn{2}{c}{2x2} & \multicolumn{2}{c}{2} & \multicolumn{2}{c}{0} \\
    \midrule     
    \end{tabular}%
  \label{tab:tab1}%
\end{table}%

To effectively extract features from both RGB and thermal images is the focus of this paper. When in the nighttime, some colorful objects in RGB maps are invisible but can be clearly seen in the thermal maps. Considering the modality difference, RGB and thermal features need to be enhanced. Inspired by \cite{woo2018cbam}, we design a FEAM module using an attention component to learn features from the fused data and then refine the prediction. In Fig.~\ref{fig:fig2}, the FEAM is added after each convolutional layer in two encoder streams, which can enhance the compatibility of the features. This extraction process improves the representation of image features and preserves the multi-level information. To better understand the working mechanism of the FEAM module, the channel-wise feature maps from FEAM are visualized at different levels.

\begin{figure*}[h]
    \includegraphics[width=6.95in]{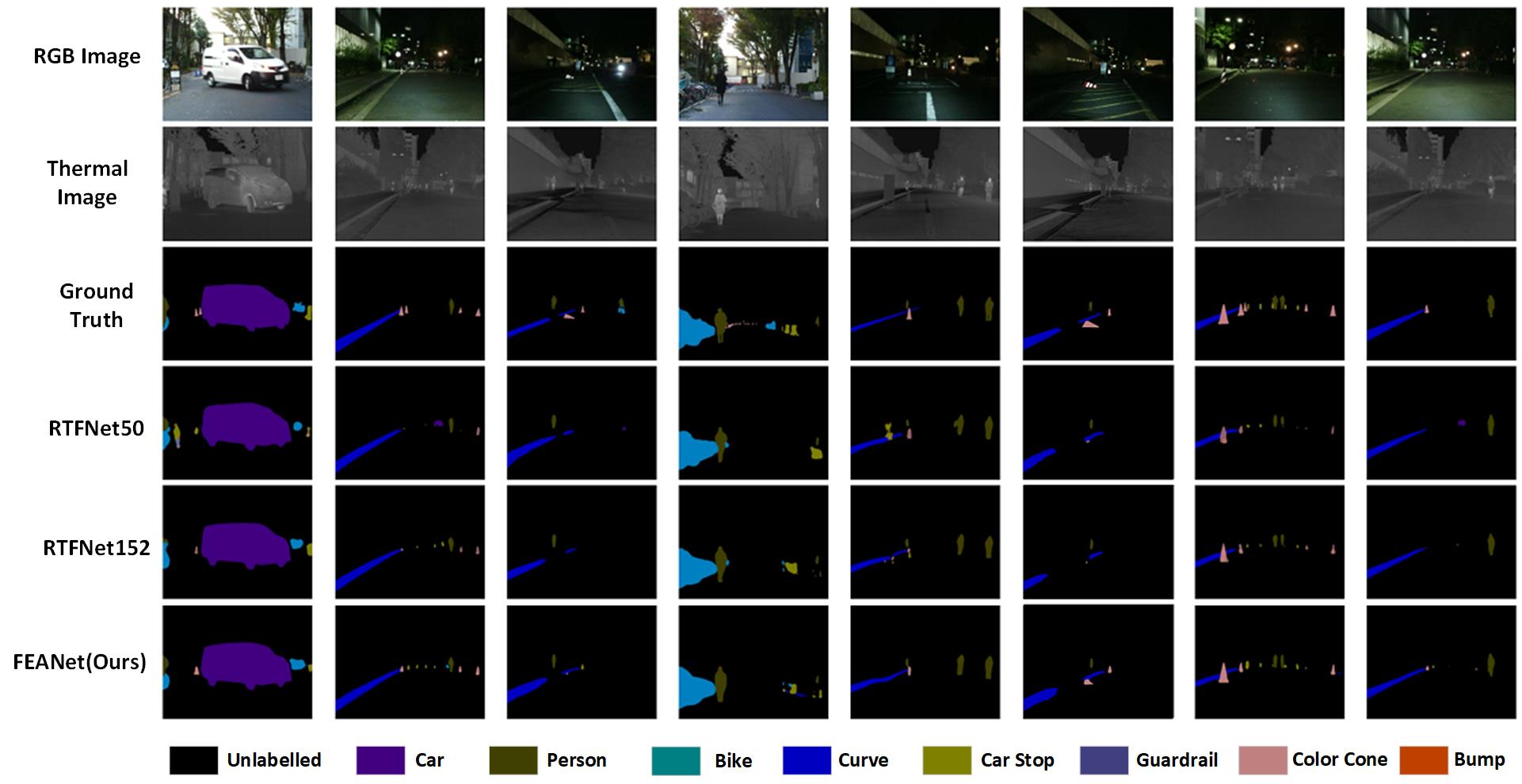}
    \vspace{-1mm}
    \caption{Qualitative demonstrations for the fusion networks in daytime or nighttime. We can see that our FEANet can provide acceptable results in various lighting conditions. The comparative results demonstrate our superiority.}
    \label{fig:fig5}
    \vspace{-5mm}
\end{figure*}

As illustrated in Fig.~\ref{fig:fig3}, the FEAM contains a sequential channel attention operation and a spatial attention operation. Channel attention operation shifts attention to the feature that extracted from the convolutional layer and then explores foreground cues. Complementarily, spatial attention operation focuses on the global area to explore the informative cues, looking for possible small target objects within it. To the best of our knowledge, we are the first to introduce the attention mechanism to excavate informative cues from both RGB and thermal multi-level features. Our experiments in Tab.~\ref{tab:tab2} and Fig.~\ref{fig:fig5} demonstrate the effectiveness of our approach in improving RGB-T semantic segmentation performance.

\subsection{Decoder-Resolution Restoring}

After computing multi-level features from two encoder streams, which are the final map of the RGB and thermal features. The decoder is mainly designed to efficiently leverage the multi-level information to carry out the detail pixels refinement. Our decoder architecture is refined from the RTFNet decoder and then restores the feature map to the original images. Different from RTFNet, we delete two sequential 1 × 1 convolutions of the original block, which avoids the complicated up-sample process in the decoder. As illustrated in Fig.~\ref{fig:fig4}, the decoder consists of two blocks (Transposed blocks A and B). Specifically, Transposed block B contains an additional branch to enlarge the receptive field and a residual connection to preserve the information. 

 Detailed configurations for the neural network layers in the Transposed blocks are displayed in Tab.~\ref{tab:tab1}. In block A, there is a batch normalization (BN) layer \cite{ioffe2015batch} and a ReLu activation layer \cite{nair2010rectified} followed by the convolutional layer as the feature resolution. The short cut from the input and the output of the final BN layer is element-wisely added up. In block B, it consists of Conv 1 and two TransConv layers. Each residual-based transposed block contains a 3×3 convolution and a residual-based transposed convolution. Through Conv 1, the resolution of the map is the same as the original one, however, the number of feature channels is decreased by a factor of 2. And the TransConv 1 keeps the number of channels unchanged and increases the resolution by a factor of 2. Different from the TransConv 1, the TransConv 2 needs to increase the resolution and decrease the number of feature channels. Finally, the decoder will get more details to generate the final predicted map in a progressive upsampling way.

\begin{table*}[htbp]
  \caption{Comparison result on the test set (\%). 3c and 4c represent that the networks are tested with the three-channel RGB data and four-channel RGB-Thermal data, respectively. Note that mAcc and mIoU are calculated with the unlabeled classes, but the results for the unlabeled classes are not displayed. The bold font highlights the best result in each column.}
  \setlength{\tabcolsep}{1.5mm}
    \begin{tabular}{cccccccccccccccccccccccc}
    \toprule
       \multicolumn{4}{c}{\multirow{2}*{Methods}}   & \multicolumn{2}{c}{Car } & \multicolumn{2}{c}{Person } & \multicolumn{2}{c}{Bike } & \multicolumn{2}{c}{Curve } & \multicolumn{2}{c}{Car Stop } & \multicolumn{2}{c}{Guardrail } & \multicolumn{2}{c}{Color Cone } & \multicolumn{2}{c}{Bump} & \multirow{2}*{mAcc}  & \multirow{2}*{mIoU} \\
       \cmidrule{5-20}    \multicolumn{4}{c}{}          & Acc   &  IoU  & Acc   &  IoU  & Acc   &  IoU  & Acc   &  IoU  & Acc   &  IoU  & Acc   &  IoU  & Acc   &  IoU  & Acc   &  IoU  &  \\
    \midrule
    \multicolumn{4}{c}{FRRN(4c)}  & 81.9  & 74.7  & 66.2  & 60.8  & 62.8  & 50.3  & 41.2  & 35.0  & 12.5  & 11.5  & 0.0   & 0.0   & 37.2  & 34.0  & 35.2  & 34.6  & 48.5  & 44.2  \\
    \multicolumn{4}{c}{FRRN(3c)}  & 80.0  & 71.2  & 53.0  & 46.1  & 65.1  & 53.0  & 34.0  & 27.1  & 21.6  & 19.1  & 0.0   & 0.0   & 34.7  & 32.5  & 36.2  & 30.5  & 47.1  & 41.8  \\
    \midrule
    \multicolumn{4}{c}{BiSeNet(4c)} & 89.7  & 84.1  & 72.0  & 63.2  & 74.1  & 60.1  & 45.1  & 36.7  & 34.2  & 25.3  & 18.2  & 5.0   & 47.4  & 42.2  & 39.8  & 35.9  & 57.7  & 50.0  \\
    \multicolumn{4}{c}{BiSeNet(3c)} & 90.0  & 84.5  & 65.0  & 54.3  & 75.0  & 61.4  & 32.1  & 25.7  & 32.3  & 26.2  & 3.2   & 0.9   & 49.6  & 43.3  & 48.1  & 40.5  & 54.9  & 48.2  \\
    \midrule
    \multicolumn{4}{c}{DFN(4c)}   & 90.0  & 84.4  & 73.2  & 65.0  & 75.5  & 60.9  & 54.0  & 40.4  & 38.9  & 25.7  & 10.2  & 2.7   & 48.3  & 42.5  & 55.8  & 47.4  & 60.5  & 52.0  \\
    \multicolumn{4}{c}{DFN(3c)}   & 90.7  & 81.4  & 67.7  & 52.8  & 71.5  & 57.5  & 49.2  & 34.9  & 35.1  & 23.8  & 4.1   & 1.4   & 44.2  & 31.0  & 54.6  & 47.5  & 57.3  & 47.5  \\
    \midrule
    \multicolumn{4}{c}{SegHRNet(4c)} & 92.8  & 87.6  & 79.3  & 71.0  & 78.3  & 63.4  & 59.8  & 42.5  & 25.7  & 19.1  & 18.8  & 0.0   & 56.5  & 49.8  & 63.5  & 44.5  & 63.7  & 53.2  \\
    \multicolumn{4}{c}{SegHRNet(3c)} & 92.2  & 86.6  & 73.1  & 59.8  & 74.9  & 61.3  & 47.0  & 33.2  & 23.8  & 28.7  & 7.3   & 0.0   & 54.6  & 47.2  & 61.5  & 46.2  & 60.9  & 51.3  \\
    \midrule
    \multicolumn{4}{c}{MFNet}     & 77.2  & 65.9  & 67.0  & 58.9  & 53.9  & 42.9  & 36.2  & 29.9  & 19.1  & 9.9   & 0.1   & 8.5   & 30.3  & 25.2  & 30.0  & 27.7  & 45.1  & 39.7  \\
    \midrule
    \multicolumn{4}{c}{FuseNet}   & 81.0  & 75.6  & 75.2  & 66.3  & 64.5  & 51.9  & 51.0  & 37.8  & 28.7  & 15.0  & 0.0   & 0.0   & 31.1  & 21.4  & 51.9  & 45.0  & 52.4  & 45.6  \\
    \midrule
    \multicolumn{4}{c}{DepthAwareCNN} & 85.2  & 77.0  & 61.7  & 53.4  & 76.0  & 56.5  & 40.2  & 30.9  & 9.9   & 29.3  & 22.8  & 6.4   & 32.9  & 30.1  & 36.5  & 32.3  & 55.1  & 46.1  \\
    \midrule
    \multicolumn{4}{c}{RTFNet-50} & 91.3  & 86.3  & 78.2  & 67.8  & 71.5  & 58.2  & 69.8  & 43.7  & 32.1  & 24.3  & 13.4  & 3.6   & 40.4  & 26.0  & 73.5  & \textbf{57.2}  & 62.2  & 51.7  \\
    \midrule
    \multicolumn{4}{c}{RTFNet-152} & 93.0  & 87.4  & 79.3  & 70.3  & 76.8  & 62.7  & 60.7  & 45.3  & \textbf{38.5 } & \textbf{29.8 } & 0.0   & 0.0   & 45.5  & 29.1  & 74.7  & 55.7  & 63.1  & 53.2  \\
    \midrule
    \multicolumn{4}{c}{FuseSeg-161} & 93.1  & \textbf{87.9}  & 81.4  & \textbf{71.7} & \textbf{78.5} & \textbf{64.6} & \textbf{68.4} & 44.8  & 29.1  & 22.7  & 63.7  & 6.4   & 55.8  & 46.9  & 66.4  & 47.9  & 70.6  & 54.5  \\
    \midrule
    \multicolumn{4}{c}{FEANet(Ours)}     & \textbf{93.3} & 87.8 & \textbf{82.7} & 71.1  & 76.7  & 61.1  & 65.5  & \textbf{46.5} & 26.6  & 22.1  & \textbf{70.8} & \textbf{6.6} & \textbf{66.6} & \textbf{55.3} & \textbf{77.3}  & 48.9  & \textbf{73.2} & \textbf{55.3} \\
    \bottomrule
    \end{tabular}%
  \label{tab:tab2}%
\end{table*}%

\section{EXPERIMENTS AND RESULTS}

\subsection{The RGB-T Dataset}

We use the public data set released by the MFNet \cite{ha2017mfnet}. It was recorded in urban street scenes, which contains eight hand-labeled object classes and one unlabelled background class. This dataset contains 1569 pairs of RGB and thermal images, in which 820 taken at daytime and 749 taken at nighttime. We follow the dataset splitting scheme proposed in \cite{ha2017mfnet}. The training set consists of 784 pairs of images. The validation set consists of 392 pairs of images. The other images are used for testing.

\subsection{Training Details}

We use the Stochastic Gradient Descent (SGD) optimization solver for training. The momentum and weight decay are set to 0.9 and 0.0005, respectively. The initial learning rate is set to 0.03. We adapt the CosineAnnealingWarmRestarts \cite{loshchilov2017decoupled} to gradually decrease the learning rate. The loss function uses the DiceLoss \cite{milletari2016v} and the SoftCrossEntropy \cite{yi2019probabilistic} for training. We give the DiceLoss and the SoftCrossEntropy a weight of 0.5 and add them to get the loss function:

$$
\textup{DiceLoss}=1-\frac{2 \sum_{i}^{N} p_{i} g_{i}}{\sum_{i}^{N} p_{i}^{2}+\sum_{i}^{N} g_{i}^{2}} \eqno{(1)}
$$
where the sums run over the $N$ voxels, of the predicted binary segmentation volume $p_{i} \in P$ and the ground truth binary volume $g_{i} \in G$. This formulation of DiceLoss can be differentiated yielding the gradient.

$$
\textup{SoftCrossEntropyloss }=-\frac{1}{n} \sum_{i=1}^{n} \sum_{j=1}^{c} \hat{y}_{i j} \log (y_{i j}^{d})\eqno{(2)}
$$
where the $n$ is the number of the batch. In this work $n = 5$, $\hat{y}_{i j}$ is is binary indicator if class label $c$ is the correct classification for pixel $i$, and $y_{i j}^{d}$ is the corresponding predicted probability be normalized to a probability distribution.

\subsection{Evaluation Metrics}
The Accuracy (Acc) and the Intersection-over-Union (IoU) are used as the evaluation indicators of our model. The Acc defines the overall accuracy as the probability of correspondence between a positive decision and true condition. The IoU calculates the intersection of the true label and the predicted result separately for each class. Both mAcc and mIoU are the average values across all the classes for the Acc and the IoU.

$$
\textup{{m}{A}{c}{c}} =\frac{1}{k+1}{\sum_{i=0}^{k}}\frac{p_{i i}}{\sum_{j=0}^{k} p_{i j}} \eqno{(3)}
$$

$$
\textup{{m}{I}{o}{U}}=\frac{1}{k+1} \sum_{i=0}^{k} \frac{p_{i i}}{\sum_{j=0}^{k} p_{i j}+\sum_{j=0}^{k} p_{j i}-p_{i i}}\eqno{(4)}
$$
where the $k$ is the number of the hand-labeled object classes, in this work, $k = 8$. $p_{i i}$ is the number of the pixels of class $i$ that are correctly classified as class $i$, $p_{i j}$ is the number of pixels of class $i$ that are wrongly classified as class $j$, $p_{j i}$ is the number of pixels of class $j$ that are wrongly classified as class $i$.

\subsection{Results And Analysis}

The complete quantitative evaluation results of the networks are listed in Tab.~\ref{tab:tab2}. Our FEANet achieves some remarkable advantages over the comparison methods in terms of both mAcc and mIoU indicators. Compared to the SOTA RGB-T semantic segmentation methods, our proposed FEANet has a great improvement on small target object detection and segmentation, especially in the Guardrail class. Compared to the SOTA FuseSeg-161, We find the Guardrail class has the  +7.1 \% Acc and +0.2 \% IoU results in improvement. At the same time, the Color Cone class has the  +5.4 \% Acc and +8.4 \% IoU results in improvement. And the other objects also have good segmentation performance. This indicates that our FEANet can make more efficient detection and segmentation on small target objects. To further demonstrate the effectiveness of our FEANet, we visualize the prediction maps of our FEANet and other top 2 methods in Fig.~\ref{fig:fig5}. Experiments show that Our FEANet effectively utilizes the RGB and thermal information for sharp object boundaries, while the others are disturbed by the background.

\begin{table}[htbp]
  \centering
  \caption{Inference speed of SOTA networks. ms and FPS represent the time of milliseconds and the speed of Frames Per Second, respectively.}
    \begin{tabular*}{\hsize}{@{}@{\extracolsep{\fill}}lllllllllllll@{}}
    \midrule
\multicolumn{3}{c}{\multirow{2}*{Methods}} & \multicolumn{3}{c}{RTX 2080 Ti} \\
\cmidrule{4-6}          &       &       & \multicolumn{1}{c}{ms} &       & \multicolumn{1}{c}{FPS} \\
    \midrule
    \multicolumn{3}{c}{RTFNet-50} & 11.25  &       & 88.87  \\
    \midrule
    \multicolumn{3}{c}{RTFNet-152} & 30.47  &       & 32.81  \\
    \midrule
    \multicolumn{3}{c}{FuseSeg-161} & 33.32  &       & 30.01  \\
    \midrule
    \multicolumn{3}{c}{FEANet(ours)} & 28.52 &       & 35.06 \\
    \bottomrule
    \end{tabular*}%
  \label{tab:tab3}%
\end{table}%

Although our FEANet can get prior results in the tiny object classes, there are some limitations in our network. According to Tab.~\ref{tab:tab2}, FuseSeg-161 gets the best results in the Person and Bike classes. This proves the effectiveness of the DenseNet, which can keep the feature map resolution unchanged in the encoder stream. And for the RTFNet, our results are better than those of the RTFNet in most object classes, however, the RTFNet152 gets the best indicators in the Car Stop class. This proves network can get the dense feature in a deeper layer to improve the segmentation performance. As indicated in Tab.~\ref{tab:tab3}, our FEANet achieves real-time inference speed (approximately 35 images/s) on a single NVIDIA Geforce RTX 2080 Ti GPU.

\subsection{Ablation Study}
\begin{table}[htbp]
  \centering
  \caption{The comparison of mAcc ($\%$) and mIoU ($\%$) on the test-sets for the NFRTS, NFRS, NFTS and FRTS(ours) variants. The bold font highlights the better results in each scenario.}
    \begin{tabular*}{\hsize}{@{}@{\extracolsep{\fill}}lllllllllllll@{}}
    \midrule
\multicolumn{3}{c}{\multirow{2}*{Variants}} & \multicolumn{3}{c}{Test-set} \\
\cmidrule{4-6}          &       &       & \multicolumn{1}{c}{mAcc} &       & \multicolumn{1}{c}{mIoU} \\
    \midrule
    \multicolumn{3}{c}{NFRTS} & 63.9  &       & 50.0  \\
    \midrule
    \multicolumn{3}{c}{NFRS} & 69.5  &       & 54.5  \\
    \midrule
    \multicolumn{3}{c}{NFTS} & 65.3  &       & 50.6  \\
    \midrule
    \multicolumn{3}{c}{FRTS(ours)} & \textbf{73.2} &       & \textbf{55.3} \\
    \bottomrule
    \end{tabular*}%
  \label{tab:tab4}%
\end{table}%
In order to verify that our FEAM module is effective at each feature level, we removed the FEAM module from the RGB stream and thermal stream, respectively, then we can see the performance without using the attention module FEAM. Therefore, we call no FEAM in a thermal stream as NFTS. Similarly, no FEAM in RGB stream as NFRS and no FEAM in RGB and thermal stream is named as NFRTS, respectively. FRTS means that the FEAM is both in RGB and thermal stream. Tab.~\ref{tab:tab4} shows the quantitative comparison test results. By comparing the results of NFRTS, NFRS, NFTS, and FRTS, we find that FRTS usually provides better performance than NFRS, NFRTS and NFTS in the RGB-T semantic segmentation task. The performance in FRTS can also prove that FEAM can enhance the fusion effect of RGB image information and thermal image information. In this experiment, the FEAM in every layer facilitates a universal improvement in detection performance. In addition, we find that FEAM applied in the thermal stream contributes more to the results.

\section{CONCLUSIONS}
We proposed a novel two-stage FEANet to excavate informative thermal cues from both RGB and thermal images for the semantic segmentation of urban scenes. Specifically, we introduce a FEAM to excavate and enhance informative features from both the channel and spatial views. The experimental results demonstrate that FEANet performs better on small target object segmentation and produces sharp object boundaries. The proposed FEANet runs at real-time speed on a single GPU, making it a potential solution for autonomous driving applications. In the future, we would like to fuse more different modalities of information (e. g., depth, audio) into a network for segmentation improvement.

\addtolength{\textheight}{-12cm}   


\bibliographystyle{IEEEtran}
\bibliography{reference.bib}

\end{document}